\documentclass{article}

\PassOptionsToPackage{numbers, compress}{natbib}

\usepackage[preprint]{nips_2018}

\usepackage[utf8]{inputenc} 
\usepackage[T1]{fontenc}    
\usepackage{hyperref}       
\usepackage{url}            
\usepackage{booktabs}       
\usepackage{amsfonts}       
\usepackage{nicefrac}       
\usepackage{microtype}      
\usepackage{amsmath}        
\usepackage{bm}             
\usepackage{epstopdf}       
\usepackage{multirow}       
\usepackage{amssymb}
\usepackage{color}
\usepackage{graphicx}

\newcommand{\method}[1]{MT-CGCNN}
\newcommand{\baseline}[1]{CGCNN}
\newcommand{\concat}{\oplus}
\newcommand{\var}{$\pm$}
\newcommand{\propone}[1]{Formation Energy}
\newcommand{\proponesym}[1]{$\Delta E^f$}
\newcommand{\proptwo}[1]{Band Gap}
\newcommand{\proptwosym}[1]{$E_g$}
\newcommand{\propthree}[1]{Fermi Energy}
\newcommand{\propthreesym}[1]{$E_F$}

\graphicspath{{plots/}} 

\title{\method{}: Integrating Crystal Graph Convolutional Neural Network with Multitask Learning for Material Property Prediction}

\author{
  Soumya Sanyal \thanks{ \,contributed equally to this paper.}\\
  Indian Institute of Science\\
  \texttt{soumyasanyal@iisc.ac.in} \\
  \And
  Janakiraman Balachandran \footnotemark[1]\\
  Shell Technology Centre Bangalore \\
  \texttt{J.Balachandran@shell.com} \\
  \And
  Naganand Yadati \\
  Indian Institute of Science \\
  \texttt{y.naganand@gmail.com} \\
  \And
  Abhishek Kumar \\
  Indian Institute of Science  \\
  \texttt{abhishekkumar12@iisc.ac.in} \\
  \And
  Padmini Rajagopalan \\
  Shell Technology Centre Bangalore \\
  \texttt{Padmini.Rajagopalan@shell.com} \\
  \And
  Suchismita Sanyal \\
  Shell Technology Centre Bangalore \\
  \texttt{Suchismita.Sanyal@shell.com} \\
  \And
  Partha Talukdar \\
  Indian Institute of Science \\
  \texttt{ppt@iisc.ac.in} \\
}

\begin{document}

\maketitle

\begin{abstract}
  Developing accurate, transferable and computationally inexpensive machine learning models can rapidly accelerate the discovery and development of new materials. Some of the major challenges involved in developing such models are, (i) limited availability of materials data as compared to other fields, (ii) lack of universal descriptor of materials to predict its various properties. The limited availability of materials data can be addressed through transfer learning, while the generic representation was recently addressed by \citet{CGCNN}, where they developed a crystal graph convolutional neural network (\baseline{}) that provides a unified representation of crystals. In this work, we develop a new model (\method{}) by integrating \baseline{}  with transfer learning based on multi-task (MT) learning. We demonstrate the effectiveness of \method{} by simultaneous prediction of various material properties such as  \propone{} (\proponesym{}), \proptwo{} (\proptwosym{}) and \propthree{} (\propthreesym{}) for a wide range of inorganic crystals (46774 materials). \method{} is able to reduce the test error when employed on correlated properties by upto 8\%. The model prediction has lower test error compared to \baseline{}, even when the training data is reduced by 10\%. We also demonstrate our model's better performance through prediction of end user scenario related to metal/non-metal classification. These results encourage further development of machine learning approaches which leverage multi-task learning to address the aforementioned challenges in the discovery of new materials. We make \method{}’s source code available to encourage reproducible research.
\end{abstract}

\section{Introduction}
The discovery, design and development of new materials with required properties underpin the development of various next generation energy, medical and electronic technologies. Discovery of new materials has historically been made through trial and error process leading to slow development cycles \cite{TomKalilMaterialsGenomeInitiative}. The advent of data driven modeling techniques has provided a new approach to develop computationally inexpensive and accurate models, that enables us to rapidly screen large material search spaces to select potential material candidates with desired properties. These approaches have recently been employed to predict new materials for various functionalities such as thermoelectrics \cite{GaultoisPerspectiveWebbasedmachine2016}, photovoltaics \cite{LuAccelerateddiscoverystable2018}, molecular light emitting diodes\cite{Gomez-BombarelliDesignefficientmolecular2016} and shape memory alloys \cite{XueAcceleratedsearchmaterials2016} among others.

One of the major challenges in developing data driven models for material discovery is the limited availability of the material datasets compared to other fields. This creates challenges in applying conventional machine learning tools for materials data. Recent works have proposed transfer learning \cite{Meredig17} and augmenting the model with pre-existing physical knowledge \cite{narendrakumarMachineLearningConstrained} to overcome this data constraint. Multi-task learning (MTL) is an important class of transfer learning algorithms that enables us to overcome such data scarcity challenges. MTL is the procedure of learning several tasks at the same time with the objective of mutually benefitting the performance of individual tasks. In this way, MTL is able to learn generalized representations (embeddings) that can explain multiple aspects of the data. Also, it is able to overcome data limitations by co-learning multiple tasks simultaneously. Using multi-task learning has shown improvements in various fields of machine learning, from natural language processing \citep{Collobert08}, computer vision \citep{Girshick15} to drug discovery \citep{Ramsundar15} and pharmaceuticals \cite{Ramsundar17} among others.

The other major challenge in material science is to be able to come up with a universal material descriptor that can be used to predict various material properties. Until recently most of the work in literature has focused on developing  hand crafted descriptors based on domain expertise \cite{HuangCommunicationUnderstandingmolecular2016,BartokGaussianapproximationpotentials2015}. However, these approaches typically are difficult to be generalized outside the tasks (properties) for which they were trained. Molecules and crystals can be defined by their chemical composition (atoms) and structure (bonding). Hence, they are naturally amenable to a generalized graph representation. Recent progress in \textit{Geometric deep learning} \citep{gdl17} has lead to formulation of graph based deep neural networks for graphical structures \citep{gnn05,gnn09,Kipf2016, gcn_iclr14}. These deep learning based approaches can automatically learn the best representation (embedding) from raw data of atoms/bonds features for different property predictions. These approaches have been successfully applied to molecules for performing various tasks such as molecular feature extraction \citep{DuvenaudMAGHAA15,gcn_camd16, mpnn_icml17} and drug discovery \citep{lddd_acs17}.  Recently, \citet{CGCNN} have developed a GCN based approach for inorganic crystals called crystal graph convolutional neural network (\baseline{}), to predict various properties of inorganic crystals.

In this work, we bridge the two approaches by augmenting \baseline{} model with multitask learning (MTL) to jointly predict multiple material properties. This approach of simultaneous prediction of different properties ensures that the generic model can automatically transfer the learning of one property to another that results in better performance. We demonstrate this approach through simultaneous prediction of various  material properties such as \propone{} (\proponesym{}), \proptwo{} (\proptwosym{}) and \propthree{} (\propthreesym{}) for a wide range of inorganic crystals (46774 materials). We also systematically explore the impact of our approach on test errors for different MTL experiments with varying amounts of training data. Finally, we also understand the impact of our method on end user scenario related to metal/non-metal classification.

\section{Background}
\label{sec:background}

\subsection{Crystal Graph Convolution Neural Network (\baseline{})}
\label{sec:cgcn}
The work by \citet{CGCNN} focuses on building a generalized crystal graph convolutional network to represent the crystals and to  predict their properties with accuracy of \textit{ab initio} physics models. A crystal graph $\mathcal{G}$ is an undirected multigraph defined by nodes representing atoms and edges representing bonds in a crystal. It allows multiple edges between the same pair of end nodes which represent the different bonds between the atoms. Thus, the graph is defined as $\mathcal{G} \mathord{=}(\mathcal{A},\mathcal{E}, \mathcal{V}, \mathcal{U})$, where $\mathcal{A}$ is the set of atoms in the crystal structure, $\mathcal{E}\mathord{=} \{(i,j)_k\mathord{:}~k^{th}~\text{bond between atoms}~i~\text{and}~j \text{ where } i,j \in \mathcal{A}\}$, is the set of undirected edges and $|\mathcal{A}|\mathord{=}N$ is the number of atoms in the crystal graph. $\bm{v_i} \in \mathcal{V}$ contains the features of the $i^{th}$ atom encoding properties of the atom. $\bm{u}_{(i, j)_k} \in \mathcal{U}$ is the feature vector for the $k^{th}$ bond between atoms $i$ and $j$. The authors propose a simple convolution function as,

\begin{equation}
\label{eq:conv_fun1}
 	\bm{v}_i^{(t+1)} = g\left[ \left( \sum_{j, k} {\bm{v}_j^{(t)} \concat \bm{u}_{(i, j)_k}} \right) \bm{W}_c^{(t)} + \bm{v}_i^{(t)} \bm{W}_s^{(t)} + \bm{b}_c^{(t)} + \bm{b}_s^{(t)} \right]
\end{equation}

where $\concat$ denotes the concatenation of atom and bond feature vectors of the neighbors of $i^{th}$ atom, $\bm{W}_c^{(t)}$, $\bm{W}_s^{(t)}$, $\bm{b}_c^{(t)}$ and $\bm{b}_s^{(t)}$ are the convolution weight matrix, self weight matrix, convolution bias and self bias of the $t$-th layer of GCN respectively, and $g(\cdot)$ is some non-linear activation function between layers.

As noted by the authors, this formulation has a shortcoming. Since the weight matrix is shared across all neighbors, equal importance is given to all the neighbors. This inherently neglects the differences of interaction strength between neighbors. To overcome this, the authors use the standard edge-gating technique \citep{MarcheggianiT17}, where the new convolution function first concatenates neighbor feature vectors $\bm{z}_{(i,j)_k}^{(t)} = \bm{v}_i^{(t)} \concat \bm{v}_j^{(t)} \concat \bm{u}_{(i, j)_k}$, and then performs convolution by,
\begin{equation}
\label{eq:conv_fun2}
	\bm{v}_i^{(t+1)} = \bm{v}_i^{(t)} + \sum_{j, k} \sigma(\bm{z}_{(i,j)_k}^{(t)} \bm{W}_c^{(t)} + \bm{b}_c^{(t)}) \odot g(\bm{z}_{(i,j)_k}^{(t)} \bm{W}_s^{(t)} + \bm{b}_s^{(t)}) 
\end{equation}
where $\odot$ denotes element-wise multiplication and $\sigma$ denotes a sigmoid function. The $\sigma(\bm{\cdot})$ acts as a learned weight matrix to incorporate different interaction strengths between neighbors.

The atom features are then pooled (using average pooling \citep{DuvenaudMAGHAA15}) to get a vector representation of the crystal $\left(\bm{v}_\mathcal{G}\right)$. This is then used as an input to a network of fully-connected layers with non-linearities which learn to predict a property value for the crystal. More concretely,
\begin{equation}
\label{eq:crystal_pooling}
	\bm{v}_\mathcal{G} = \frac{1}{N}\sum_{i} \bm{v}_i
\end{equation}
\begin{equation}
\label{eq:fc_layer}
	\hat{y} = f\left( \bm{v}_\mathcal{G} \bm{W}_g + \bm{b}_g \right)
\end{equation}
where $\bm{v}_i$ is the learned feature representation of $i^{th}$ atom using Eq. \ref{eq:conv_fun2}, $\bm{v}_\mathcal{G}$ is the crystal representation learned from pooling and $\hat{y}$ is the predicted value of the crystal property. $\bm{W}_g$, $\bm{b}_g$ and $f(\cdot)$ are the weight matrix, bias and non-linearities of the fully-connected network respectively.

\subsection{Multi-task learning}
\label{sec:mtl_background}
The fundamental motivation for doing multi-task learning is to achieve better generalization performance. As summarized by \citep{Caruana1997}, "MTL improves generalization by leveraging the domain-specific information contained in the training signals of \textit{related} tasks". The two main architectures for MTL in the deep learning context \citep{Ruder17a} are:

\begin{itemize}
\item \textbf{Hard parameter sharing:} This is the simplest approach to MTL. The architecture shares a common set of layers across all tasks and then some task-specific output layers are present for each individual task. The key motivation is to force the model to learn better representations that can be used to learn multiple related tasks at the same time.

\item \textbf{Soft parameter sharing:} Here, there are independent models with own set of parameters for each of the tasks being learned. But then, the distance between the parameters ($l_2$ distance) are regularized to encourage learning of similar parameters for the different models. This indirectly leads to a generalized representation with the flexibility of unique parameters for each task.
\end{itemize}

A more detailed discussion on various aspects of multi-task learning could be found in \citep{Caruana1997, Ruder17a}

\section{Proposed method (MT-CGCNN)}
\label{sec:method}

\begin{figure}[htp]

\centering
\includegraphics[width=\linewidth]{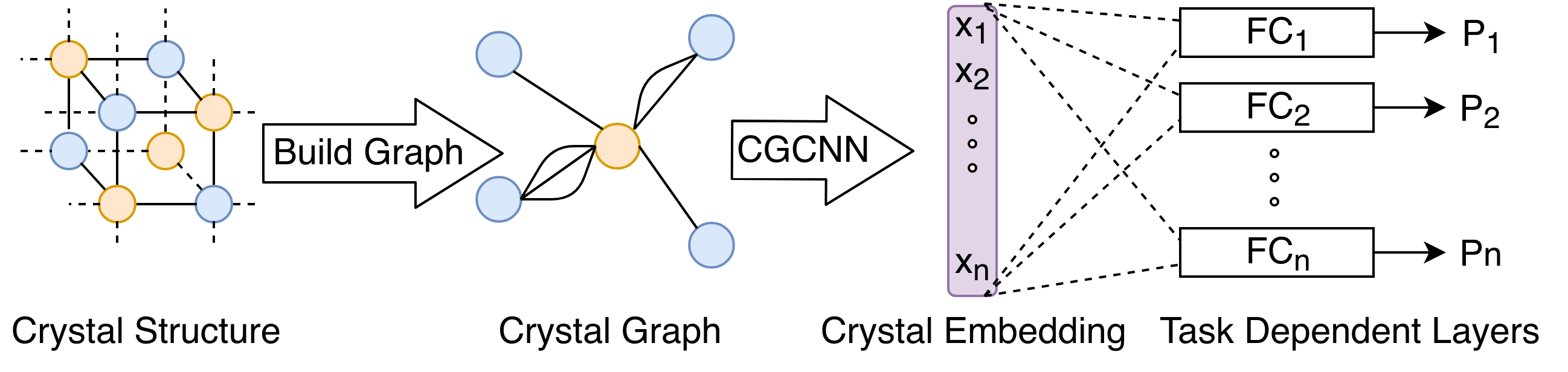}

\caption{(best viewed in color) Overview of \method{}: Given a crystal structure, a crystal graph is created from it. Note that the graph created can have multiple edges between the atoms representing different atomic bonds. Next, \baseline{} is used to extract the crystal representation using Graph Convolutional Networks. The crystal representation is then used as input for different task-specific fully connected layers ($FC_n$) which predict some property of the crystal. Refer to section \ref{sec:method} for more details.}
\label{fig:method}

\end{figure}

Fig. \ref{fig:method} shows the schematics of the MT-CGCNN model setup. Every atom and bond between atoms in a crystal has some initial vector representation \citep{CGCNN}. The feature embedding for atoms ($\bm{v}_i$) and bonds ($\bm{u}_{(i, j)_k}$) are the input to the GCN layers. Stacked GCN layers are used to encode these atomic representations using Eq. \ref{eq:conv_fun2}. This is then followed by a pooling layer (Eq. \ref{eq:crystal_pooling}) which gives a vector representation for the crystal structure $\bm{v}_\mathcal{G}$. We then use \textit{hard parameter sharing} MTL, where for each crystal property $\left(p\right)$ being learned, there is an independent fully-connected network which takes $\bm{v}_\mathcal{G}$ and predicts the property value as,

\begin{equation}
\label{eq:mtl}
	\widehat{y_{p}} = f_{p}\left( \bm{v}_\mathcal{G} \bm{W}_{p} + \bm{b}_{p} \right)
\end{equation}
      
where $\widehat{y_{p}}$ is the crystal property value for the $p^{th}$ property. $\bm{W}_{p}$, $\bm{b}_{p}$ and $f_{p}(\cdot)$ are the weight matrix, bias and non-linear mapping of the $p^{th}$ fully-connected network respectively. So, each task essentially shares the crystal representation $\bm{v}_\mathcal{G}$ and tries to learn functions that can predict a set of crystal properties. In this work, we employ mean squared loss function for each property. The total loss function for the network is the weighted linear sum of individual losses from parts of the network. This formulation of the total loss function is a common setup for the multi-tasking problem \citep{Gradnorm17, KendallGC17}. Mathematically,

\begin{equation}
\label{eq:mtl_loss}
	\mathcal{L} = \frac{1}{|\mathcal{P}|} \sum_{p \in \mathcal{P}} w_{p}L_{p}
      \end{equation}
      
where $\mathcal{L}$ is the total loss of the network, $L_{p}$ are individual losses from each of the task-specific layers and $w_{p}$ are the weights for the individual losses. A trivial setup is where $w_{p}\mathord{=}1$ which gives an average loss across tasks. For our experiments, each of $L_{p}$ is mean squared error defined by
\begin{equation}
\label{eq:mse_loss}
	L_{p} = \frac{1}{batchsize} \sum_{p \in \mathcal{P}} \left( \widehat{y_{p}} - y_{p} \right)^2
\end{equation}
where $batchsize$ is the mini-batch size during an iteration. $\widehat{y_{p}}$ is the model predicted property value and $y_{p}$ is the target property value for the $p^{th}$ property. Finally, back-propagation using gradient descent \citep{Rumelhart1988} is done to train the model. The source code for \method{} is available at \href{https://github.com/soumyasanyal/mt-cgcnn}{https://github.com/soumyasanyal/mt-cgcnn}.

\section{Experiments and results}

\subsection{Dataset}
\label{sec:dataset}
MT-CGCNN is trained and validated on inorganic crystal data comprising of 46774 materials used by \citet{CGCNN} which is obtained from the Materials Project (MP) \citep{Jain2013}. In our experiments, we focus on three correlated properties namely, \propone{} (\proponesym{}), \proptwo{} (\proptwosym{}) and \propthree{} (\propthreesym{}).

\subsection{Correlation between properties}
\label{sec:correlation}
\begin{figure}[htp]

\centering
\includegraphics[width=\textwidth]{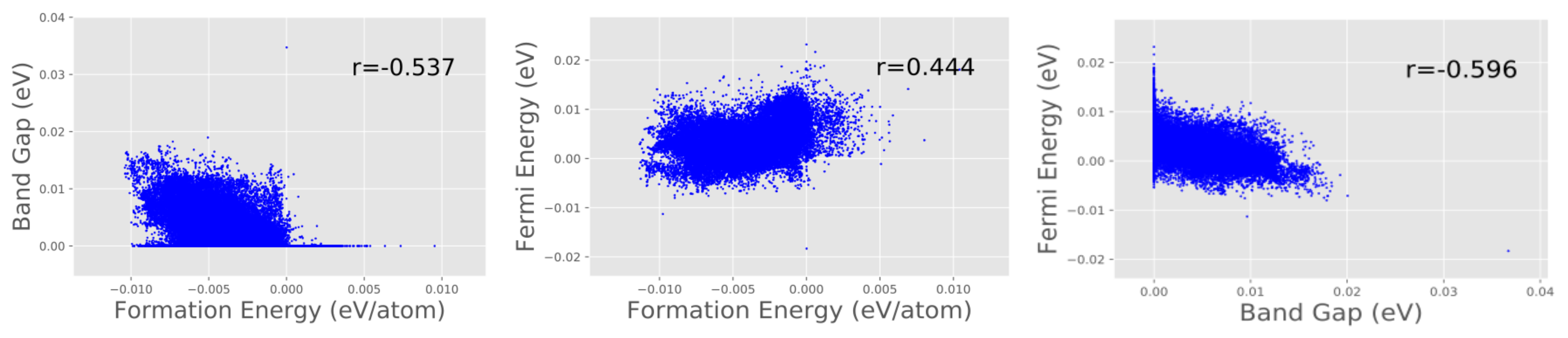}\hfill
\caption{Correlation plots between different properties.}
\label{fig:correlation}

\end{figure}

One of the crucial problems in multitasking is to understand which tasks could probably help in an MTL setup \cite{Caruana1997, Ruder17a}. While there have been advancements towards understanding that problem \cite{Xu17, Bingel17}, in our setup we select tasks which have significant correlation. The Pearson correlation coefficients \citep{benesty2009pearson} for the three properties -- \proponesym{}, \proptwosym{} and \propthreesym{} are shown in Fig. \ref{fig:correlation}.

\subsection{Weighted loss}
Weighted loss as defined in Eq. \ref{eq:mtl_loss} is useful for cases when we want to give more importance to one task over another. This may be needed in cases when a specific task is harder to learn than the rest and hence would not get equally trained as others \cite{Gradnorm17}. In our current setup, we consider these weights as hyperparameters for the model and search for the best weights.

\subsection{Model evaluation}
\label{sec:results}
To evaluate \method{}, we run a set of experiments with setup as detailed in Table \ref{table:experiments}. The results from our experiments are summarized in Table \ref{table:avg} and Table \ref{table:results}. We report mean absolute error (MAE) over 5 runs with random splits of 60/20/20 ratio of train, validation and test sets, unless specified otherwise. To get the numbers for the \baseline{} model, we used the code provided by the authors \footnote{https://github.com/txie-93/cgcnn} with the hyperparameters reported in their work.

\begin{table}[h]
	\caption{Experimental Setup for evaluation}
	\label{table:experiments}
	\centering
	\begin{tabular}{ll}
		\toprule
		Experiment 			&Setup 											\\
		\midrule
		E1					&\propone{} (\proponesym{}) and \proptwo{} (\proptwosym{})				\\
		E2					&\propone{} (\proponesym{}) and \propthree{} (\propthreesym{})				\\
		E3					&\proptwo{} (\proptwosym{}) and \propthree{} (\propthreesym{}) 					\\
		E4					&\propone{} (\proponesym{}), \proptwo{} (\proptwosym{}) and \propthree{} (\propthreesym{})	\\
		\bottomrule
	\end{tabular}
\end{table}

In Table \ref{table:avg}, the average MAE (the average of MAEs for individual properties) is tabulated with the relative increase in performance over the baseline due to multi-tasking. Here, we can see that multi-task learning clearly outperforms the single-task \baseline{} model across all the experiments. In Table \ref{table:results} we show how our model performs on individual properties compared to single task setup (\baseline{}). For example, we observe a strong reduction in the MAE scores of \proptwosym{} when we do multi-tasking using \proptwosym{} and \proponesym{}. A similar trend is observed for \propthreesym{} when we do multi-tasking using \proponesym{} and \propthreesym{}. These observations indicate that multi-tasking is more helpful when done with a specific combination of tasks. We observe from Table \ref{table:results} that \proponesym{} prediction shows degradation during multi-task learning, likely due to the strong constraints of hard parameter sharing.

Further, we do another set of experiment where we systematically reduce the training data available to the different models and check the model performance for the reduced training dataset. The results are shown in Table \ref{table:reduced}. We observe that \method{} outperforms \baseline{} for the same amount of input data. Specifically, we note that the MAE values of \method{} using 50\% training data is better than \baseline{} using 60\% training data. This is a reduction of approximately 4.5k training samples for the current setup. This result verifies that multi-tasking leads to comparable performance even with lesser training data. Also, it indirectly shows that multi-tasking leads to a faster learning of the crystal embedding space.

\begin{table}[h]
	\caption{Average MAE values with percentage of improvement for different experiments on \proponesym{}, \proptwosym{} and \propthreesym{}. Our model performs consistently better than baseline (\baseline{}). Refer section \ref{sec:results} for more details.}
	\label{table:avg}
	\centering
	\begin{tabular}{llll}
		\toprule
		Experiment 			&\baseline{}	&\method{}		&Improvement(\%)	\\
		\midrule
		E1					& 0.181			&\bf{0.166}		& 8.3\%				\\
		E2					& 0.210			&\bf{0.202}		& 3.8\%				\\
		E3					& 0.352			&\bf{0.346}		& 1.7\%				\\
		E4					& 0.247			&\bf{0.236}		& 4.4\%				\\
		\bottomrule
	\end{tabular}
\end{table}

\begin{table}[h]
	\caption{Individual MAE of three properties - \proponesym{}, \proptwosym{} and \propthreesym{} using \baseline{} and \method{} models. Our model performs better for \proptwosym{} and \propthreesym{} prediction. Refer section \ref{sec:results} for more details.}
	\label{table:results}
	\centering
	\begin{tabular}{lllll}
		\toprule
		Method 							&Experiment			&\proponesym{} (eV/atom)&\proptwosym{} (eV)		&\propthreesym{} (eV)	\\
		\midrule
		\multirow{3}{*}{\baseline{}}	&\proponesym{}		&\bf{0.039} \var 0.0003	&-						&-						\\
										&\proptwosym{}		&-						&0.323 \var 0.006		&-						\\
										&\propthreesym{}	&-						&-						&0.380 \var 0.006		\\
		\midrule
		\multirow{4}{*}{\method{}}		&E1					& 0.043 \var 0.001		&\bf{0.290} \var 0.004	&-						\\
										&E2					& 0.041	\var 0.001		&-						&\bf{0.363} \var 0.003	\\
										&E3					& -						&0.319 \var 0.004		&0.373 \var 0.003		\\
										&E4					& 0.050 \var 0.002		&0.295 \var 0.004		&\bf{0.363} \var 0.006	\\
		\bottomrule
	\end{tabular}
\end{table}

\begin{table}[h]
	\caption{MAE values of \proponesym{} and \proptwosym{} with increasing training data split from 20\% to 60\%. Our model performs better with 50\% training data compared to baseline with 60\% training data (highlighted in bold). Refer section \ref{sec:results} for more details.}
	\label{table:reduced}
	\centering
	\begin{tabular}{lllllllllll}
		\toprule
		\multirow{2}{*}{Property}	&\multicolumn{5}{c}{\baseline{}}		&\multicolumn{5}{c}{\method{}}		\\
									\cmidrule(r){2-6}						\cmidrule(r){7-11}
									&20\%&30\%&40\%&50\%&60\%				&20\%&30\%&40\%&50\%&60\%			\\
		\midrule
		\proponesym{}				&0.062&0.052&0.046&0.043&0.039			&0.062&0.053&0.049&0.046&0.043		\\
		\proptwosym{}				&0.424&0.385&0.356&0.332&0.323			&0.388&0.346&0.326&0.301&0.290		\\
		\midrule
		Avg MAE 					&0.243&0.218&0.201&0.188&\bf{0.181}		&0.225&0.200&0.188&\bf{0.174}&0.166	\\
		\bottomrule
	\end{tabular}
\end{table}

\subsection{End user scenarios (chemical insights)}
\label{sec:Insights}

Beyond test error evaluation, we also evaluate our model on scenarios that are useful for the end users. In the case of material scientists and chemists, this translates into obtaining chemical insights from the predicted data. This, in turn, provides another framework to compare the two approaches. Here, we analyze two scenarios that can provide some chemical insights.

\begin{figure}[htp]
\centering
\includegraphics[width=\linewidth]{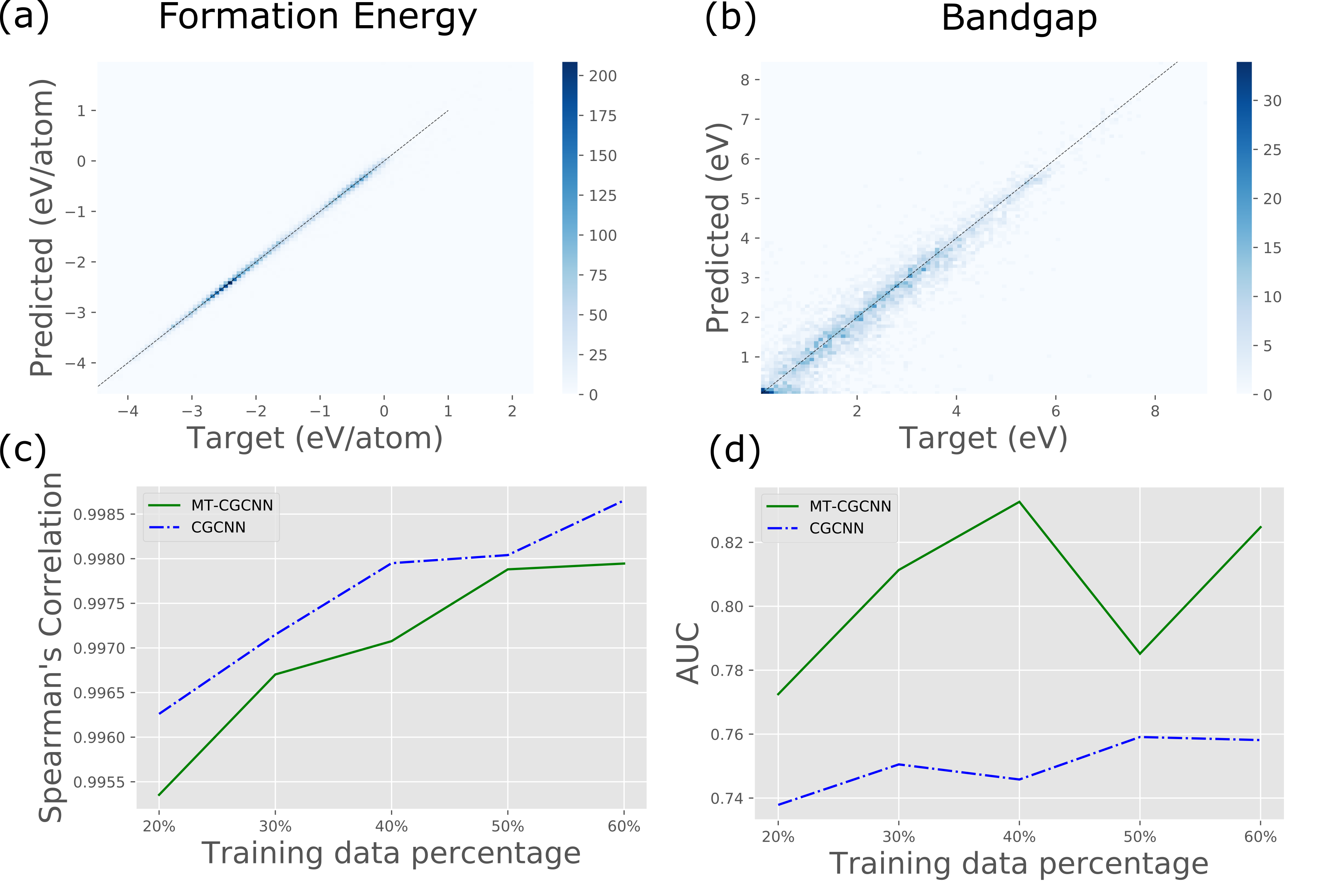}\hfill
\caption{(best viewed in color) (a) Predicted \proponesym{} (vs) true \proponesym{} for 60\% training data. (b) Predicted \proptwosym{} (vs) true \proptwosym{} for 60\% training data. (c) Spearman's rank correlation coefficient  $\left(r_s\right)$ of predicted \proponesym{} and true \proponesym{} for \method{} and \baseline{} as a function of training data. Our model is comparable with the baseline. (d) Area under the curve (AUC) of metal/non-metal classification for \method{} and \baseline{} as a function of training data. The lowest AUC of our model is higher than the highest AUC of the baseline. Refer section \ref{sec:Insights} for more details.}
\label{fig:reduced_data}
\end{figure}

For the first scenario, we compare the ordering of different materials based on Formation energy. The difference between Formation energy helps to understand the relative stability of different materials. Hence, from the end user standpoint, it is more important to rank the crystals correctly using the \proponesym{} rather than the accuracy of prediction. To quantify this ordering (ranking) of materials, we calculate the Spearman's rank correlation coefficient $\left(r_s\right)$ \cite{myers2003research} for the predicted \proponesym{} and true \proponesym{} using \method{} and \baseline{} for different amounts of training data as shown in Fig. \ref{fig:reduced_data}(c). The $r_s$ values of both the approaches are very high and comparable. This suggests that the ordering between the crystals based on their \proponesym{} is mostly preserved.

In case of second scenario, based on \proptwosym{} we classify the materials into two classes namely (i) \textit{metals} -- that can easily conduct electrons and (ii) \textit{non-metals} such as semiconductors and insulators where electron conduction is constrained. The energy equivalent of a physical system maintained at temperature $T$ is calculated as $k_BT$, where $k_B$ is Boltzmann constant. In case of room temperature ($T=300K$), this value is ~0.025eV. Hence, crystals with \proptwosym{} less than 0.025 eV are considered metals, while the rest of them are considered non-metals comprising of semiconductors and insulators. Fig. \ref{fig:reduced_data}(d) shows the area under the curve (AUC) for crystal classification into metal/non-metal using \method{} and \baseline{} for different amounts of training data. It can be observed that \method{} has a much higher accuracy in classification compared to \baseline{} as measured by the AUC metric. In fact, as a function of training data, the lowest AUC of \method{} is still higher than the highest AUC of \baseline{}.

\subsection{Hyperparameters}
\label{sec:hyperparams}
We divide the dataset into train, validation and test splits. To tune the hyperparameters, we train the model using the training set and then check the test error on the validation set. We perform grid search with early stopping over the hyperparameter space mentioned in Table \ref{table:hyperparam}. For training, we use Adam optimizer \citep{KingmaB14} with a learning rate of 0.01.

\begin{table}[h]
	\caption{A list of hyperparameters with values on which grid search is performed}
	\label{table:hyperparam}
	\centering
	\begin{tabular}{ll}
		\toprule
		Hyperparameter 												& Values 										\\
		\midrule
		Number of convolutional layers 								& 1, 2, 3, 4, 5    								\\
		Length of learned atom feature vector $\bm{v}_i$ 			& 16, 32, 64, 128 								\\
		Length of graph hidden representation						& 16, 32, 64, 128								\\
		Number of hidden fully-connected layers per task 			& 1, 2, 3, 4 									\\
		$L_2$ Regularization term 									& $0, 10^{-6}, 10^{-4}$ 						\\
		Step size of the Adam optimizer 							& $10^{-4}$, $10^{-3}$, $10^{-2}$, $10^{-1}$ 	\\
		Weights in the weighted loss (Eq. \ref{eq:mtl_loss})		& 1, 2, 3, 4, 5, 6, 7							\\
		\bottomrule
	\end{tabular}
\end{table}

\hfill\break
\hfill\break
\hfill\break
\hfill\break
\hfill\break
\hfill\break
\hfill\break

\section{Conclusion}
\label{sec:conclusion}
In summary, we propose \method{}, an effective multi-tasking framework that uses crystal graph convolutions to predict different material properties (\proponesym{}, \proptwosym{}, \propthreesym{}) by exploiting the correlation between them. We also show that \method{} can achieve comparable accuracy as \baseline{} with fewer training samples. Additionally, we demonstrate the effectiveness of \method{} by testing some end user scenarios relating to the ordering of crystal based on \proponesym{} and classification of materials based on \proptwosym{}. The ability to predict multiple properties shows that the material representation learned is well generalized. This work opens up new research directions for machine learning with material science, where we can continue to build upon the framework of \method{} (eg. including soft-parameter sharing) to predict other functional properties of materials with limited input data. Also, exploring dynamic weighted loss has the advantage of not requiring extensive hyperparameter tuning. Integrating this with \method{} is left for future works \citep{Gradnorm17, KendallGC17}. We make \method{}’s source code available to encourage reproducible research \footnote{https://github.com/soumyasanyal/mt-cgcnn}.

\subsubsection*{Acknowledgments}
This work was funded by Shell. We would like to thank Professor Umesh Waghmare from Jawaharlal Nehru Centre for Advanced Scientific Research and Professor Arnab Bhattacharyya from Indian Institute of Science for their insightful discussions. We would also like to thank Tian Xie for providing clarifications on various aspects of the \baseline{} code.

\bibliographystyle{achemso}
\bibliography{references}

\providecommand{\latin}[1]{#1}
\makeatletter
\providecommand{\doi}
  {\begingroup\let\do\@makeother\dospecials
  \catcode`\{=1 \catcode`\}=2 \doi@aux}
\providecommand{\doi@aux}[1]{\endgroup\texttt{#1}}
\makeatother
\providecommand*\mcitethebibliography{\thebibliography}
\csname @ifundefined\endcsname{endmcitethebibliography}
  {\let\endmcitethebibliography\endthebibliography}{}
\begin{mcitethebibliography}{35}
\providecommand*\natexlab[1]{#1}
\providecommand*\mciteSetBstSublistMode[1]{}
\providecommand*\mciteSetBstMaxWidthForm[2]{}
\providecommand*\mciteBstWouldAddEndPuncttrue
  {\def\EndOfBibitem{\unskip.}}
\providecommand*\mciteBstWouldAddEndPunctfalse
  {\let\EndOfBibitem\relax}
\providecommand*\mciteSetBstMidEndSepPunct[3]{}
\providecommand*\mciteSetBstSublistLabelBeginEnd[3]{}
\providecommand*\EndOfBibitem{}
\mciteSetBstSublistMode{f}
\mciteSetBstMaxWidthForm{subitem}{(\alph{mcitesubitemcount})}
\mciteSetBstSublistLabelBeginEnd
  {\mcitemaxwidthsubitemform\space}
  {\relax}
  {\relax}

\bibitem[Xie and Grossman(2018)Xie, and Grossman]{CGCNN}
Xie,~T.; Grossman,~J.~C. Crystal Graph Convolutional Neural Networks for an
  Accurate and Interpretable Prediction of Material Properties. \emph{Phys.
  Rev. Lett.} \textbf{2018}, \emph{120}, 145301\relax
\mciteBstWouldAddEndPuncttrue
\mciteSetBstMidEndSepPunct{\mcitedefaultmidpunct}
{\mcitedefaultendpunct}{\mcitedefaultseppunct}\relax
\EndOfBibitem
\bibitem[{Tom Kalil} and {Cyrus Wadia}(){Tom Kalil}, and {Cyrus
  Wadia}]{TomKalilMaterialsGenomeInitiative}
{Tom Kalil},; {Cyrus Wadia}, \emph{Materials {{Genome Initiative}} for {{Global
  Competitiveness}}}\relax
\mciteBstWouldAddEndPuncttrue
\mciteSetBstMidEndSepPunct{\mcitedefaultmidpunct}
{\mcitedefaultendpunct}{\mcitedefaultseppunct}\relax
\EndOfBibitem
\bibitem[Gaultois \latin{et~al.}(2016)Gaultois, Oliynyk, Mar, Sparks,
  Mulholland, and Meredig]{GaultoisPerspectiveWebbasedmachine2016}
Gaultois,~M.~W.; Oliynyk,~A.~O.; Mar,~A.; Sparks,~T.~D.; Mulholland,~G.~J.;
  Meredig,~B. Perspective: {{Web}}-Based Machine Learning Models for Real-Time
  Screening of Thermoelectric Materials Properties. \emph{APL Materials}
  \textbf{2016}, \emph{4}, 053213\relax
\mciteBstWouldAddEndPuncttrue
\mciteSetBstMidEndSepPunct{\mcitedefaultmidpunct}
{\mcitedefaultendpunct}{\mcitedefaultseppunct}\relax
\EndOfBibitem
\bibitem[Lu \latin{et~al.}(2018)Lu, Zhou, Ouyang, Guo, Li, and
  Wang]{LuAccelerateddiscoverystable2018}
Lu,~S.; Zhou,~Q.; Ouyang,~Y.; Guo,~Y.; Li,~Q.; Wang,~J. Accelerated Discovery
  of Stable Lead-Free Hybrid Organic-Inorganic Perovskites via Machine
  Learning. \emph{Nature Communications} \textbf{2018}, \emph{9}\relax
\mciteBstWouldAddEndPuncttrue
\mciteSetBstMidEndSepPunct{\mcitedefaultmidpunct}
{\mcitedefaultendpunct}{\mcitedefaultseppunct}\relax
\EndOfBibitem
\bibitem[{G\'omez-Bombarelli} \latin{et~al.}(2016){G\'omez-Bombarelli},
  {Aguilera-Iparraguirre}, Hirzel, Duvenaud, Maclaurin, {Blood-Forsythe}, Chae,
  Einzinger, Ha, Wu, Markopoulos, Jeon, Kang, Miyazaki, Numata, Kim, Huang,
  Hong, Baldo, Adams, and
  {Aspuru-Guzik}]{Gomez-BombarelliDesignefficientmolecular2016}
{G\'omez-Bombarelli},~R. \latin{et~al.}  Design of Efficient Molecular Organic
  Light-Emitting Diodes by a High-Throughput Virtual Screening and Experimental
  Approach. \emph{Nature Materials} \textbf{2016}, \emph{15}, 1120--1127\relax
\mciteBstWouldAddEndPuncttrue
\mciteSetBstMidEndSepPunct{\mcitedefaultmidpunct}
{\mcitedefaultendpunct}{\mcitedefaultseppunct}\relax
\EndOfBibitem
\bibitem[Xue \latin{et~al.}(2016)Xue, Balachandran, Hogden, Theiler, Xue, and
  Lookman]{XueAcceleratedsearchmaterials2016}
Xue,~D.; Balachandran,~P.~V.; Hogden,~J.; Theiler,~J.; Xue,~D.; Lookman,~T.
  Accelerated Search for Materials with Targeted Properties by Adaptive Design.
  \emph{Nature Communications} \textbf{2016}, \emph{7}, 11241\relax
\mciteBstWouldAddEndPuncttrue
\mciteSetBstMidEndSepPunct{\mcitedefaultmidpunct}
{\mcitedefaultendpunct}{\mcitedefaultseppunct}\relax
\EndOfBibitem
\bibitem[Hutchinson \latin{et~al.}(2017)Hutchinson, Antono, Gibbons, Paradiso,
  Ling, and Meredig]{Meredig17}
Hutchinson,~M.~L.; Antono,~E.; Gibbons,~B.~M.; Paradiso,~S.; Ling,~J.;
  Meredig,~B. Overcoming data scarcity with transfer learning. \emph{CoRR}
  \textbf{2017}, \emph{abs/1711.05099}\relax
\mciteBstWouldAddEndPuncttrue
\mciteSetBstMidEndSepPunct{\mcitedefaultmidpunct}
{\mcitedefaultendpunct}{\mcitedefaultseppunct}\relax
\EndOfBibitem
\bibitem[{Narendra Kumar} \latin{et~al.}(){Narendra Kumar}, {Padmini
  Rajagopalan}, {Praveen Pankajakshan}, {Arnab Bhattacharyya}, {Suchismita
  Sanyal}, {Janakiraman Balachandran}, and {Umesh V.
  Waghmare}]{narendrakumarMachineLearningConstrained}
{Narendra Kumar},; {Padmini Rajagopalan},; {Praveen Pankajakshan},; {Arnab
  Bhattacharyya},; {Suchismita Sanyal},; {Janakiraman Balachandran},; {Umesh V.
  Waghmare}, Machine Learning Constrained with Dimensional Analysis and Scaling
  Laws: {{Simple}}, Transferable and Interpretable Models of Materials from
  Small Datasets. \emph{(in review)} \relax
\mciteBstWouldAddEndPunctfalse
\mciteSetBstMidEndSepPunct{\mcitedefaultmidpunct}
{}{\mcitedefaultseppunct}\relax
\EndOfBibitem
\bibitem[Collobert and Weston(2008)Collobert, and Weston]{Collobert08}
Collobert,~R.; Weston,~J. A Unified Architecture for Natural Language
  Processing: Deep Neural Networks with Multitask Learning. Proceedings of the
  25th International Conference on Machine Learning. New York, NY, USA, 2008;
  pp 160--167\relax
\mciteBstWouldAddEndPuncttrue
\mciteSetBstMidEndSepPunct{\mcitedefaultmidpunct}
{\mcitedefaultendpunct}{\mcitedefaultseppunct}\relax
\EndOfBibitem
\bibitem[Girshick(2015)]{Girshick15}
Girshick,~R.~B. Fast {R-CNN}. 2015 {IEEE} International Conference on Computer
  Vision, {ICCV} 2015, Santiago, Chile, December 7-13, 2015. 2015; pp
  1440--1448\relax
\mciteBstWouldAddEndPuncttrue
\mciteSetBstMidEndSepPunct{\mcitedefaultmidpunct}
{\mcitedefaultendpunct}{\mcitedefaultseppunct}\relax
\EndOfBibitem
\bibitem[{Ramsundar} \latin{et~al.}(2015){Ramsundar}, {Kearnes}, {Riley},
  {Webster}, {Konerding}, and {Pande}]{Ramsundar15}
{Ramsundar},~B.; {Kearnes},~S.; {Riley},~P.; {Webster},~D.; {Konerding},~D.;
  {Pande},~V. {Massively Multitask Networks for Drug Discovery}. \emph{ArXiv
  e-prints} \textbf{2015}, \relax
\mciteBstWouldAddEndPunctfalse
\mciteSetBstMidEndSepPunct{\mcitedefaultmidpunct}
{}{\mcitedefaultseppunct}\relax
\EndOfBibitem
\bibitem[Ramsundar \latin{et~al.}(2017)Ramsundar, Liu, Wu, Verras, Tudor,
  Sheridan, and Pande]{Ramsundar17}
Ramsundar,~B.; Liu,~B.; Wu,~Z.; Verras,~A.; Tudor,~M.; Sheridan,~R.~P.;
  Pande,~V. Is Multitask Deep Learning Practical for Pharma? \emph{Journal of
  Chemical Information and Modeling} \textbf{2017}, \emph{57}, 2068--2076,
  PMID: 28692267\relax
\mciteBstWouldAddEndPuncttrue
\mciteSetBstMidEndSepPunct{\mcitedefaultmidpunct}
{\mcitedefaultendpunct}{\mcitedefaultseppunct}\relax
\EndOfBibitem
\bibitem[Huang and {von Lilienfeld}(2016)Huang, and {von
  Lilienfeld}]{HuangCommunicationUnderstandingmolecular2016}
Huang,~B.; {von Lilienfeld},~O.~A. Communication: {{Understanding}} Molecular
  Representations in Machine Learning: {{The}} Role of Uniqueness and Target
  Similarity. \emph{The Journal of Chemical Physics} \textbf{2016}, \emph{145},
  161102\relax
\mciteBstWouldAddEndPuncttrue
\mciteSetBstMidEndSepPunct{\mcitedefaultmidpunct}
{\mcitedefaultendpunct}{\mcitedefaultseppunct}\relax
\EndOfBibitem
\bibitem[Bart\'ok and Cs\'anyi(2015)Bart\'ok, and
  Cs\'anyi]{BartokGaussianapproximationpotentials2015}
Bart\'ok,~A.~P.; Cs\'anyi,~G. Gaussian Approximation Potentials: {{A}} Brief
  Tutorial Introduction. \emph{International Journal of Quantum Chemistry}
  \textbf{2015}, \emph{115}, 1051--1057\relax
\mciteBstWouldAddEndPuncttrue
\mciteSetBstMidEndSepPunct{\mcitedefaultmidpunct}
{\mcitedefaultendpunct}{\mcitedefaultseppunct}\relax
\EndOfBibitem
\bibitem[Bronstein \latin{et~al.}(2017)Bronstein, Bruna, LeCun, Szlam, and
  Vandergheynst]{gdl17}
Bronstein,~M.~M.; Bruna,~J.; LeCun,~Y.; Szlam,~A.; Vandergheynst,~P. Geometric
  Deep Learning: Going beyond Euclidean data. \emph{{IEEE} Signal Process.
  Mag.} \textbf{2017}, \relax
\mciteBstWouldAddEndPunctfalse
\mciteSetBstMidEndSepPunct{\mcitedefaultmidpunct}
{}{\mcitedefaultseppunct}\relax
\EndOfBibitem
\bibitem[Gori \latin{et~al.}(2005)Gori, Monfardini, and Scarselli]{gnn05}
Gori,~M.; Monfardini,~G.; Scarselli,~F. A new model for learning in graph
  domains. Proceedings. 2005 IEEE International Joint Conference on Neural
  Networks (IJCNN). 2005; pp 729--734\relax
\mciteBstWouldAddEndPuncttrue
\mciteSetBstMidEndSepPunct{\mcitedefaultmidpunct}
{\mcitedefaultendpunct}{\mcitedefaultseppunct}\relax
\EndOfBibitem
\bibitem[Scarselli \latin{et~al.}(2009)Scarselli, Gori, Tsoi, Hagenbuchner, and
  Monfardini]{gnn09}
Scarselli,~F.; Gori,~M.; Tsoi,~A.~C.; Hagenbuchner,~M.; Monfardini,~G. The
  Graph Neural Network Model. \emph{Trans. Neur. Netw.} \textbf{2009},
  \emph{20}, 61--80\relax
\mciteBstWouldAddEndPuncttrue
\mciteSetBstMidEndSepPunct{\mcitedefaultmidpunct}
{\mcitedefaultendpunct}{\mcitedefaultseppunct}\relax
\EndOfBibitem
\bibitem[Kipf and Welling(2017)Kipf, and Welling]{Kipf2016}
Kipf,~T.~N.; Welling,~M. Semi-Supervised Classification with Graph
  Convolutional Networks. International Conference on Learning Representations
  (ICLR). 2017\relax
\mciteBstWouldAddEndPuncttrue
\mciteSetBstMidEndSepPunct{\mcitedefaultmidpunct}
{\mcitedefaultendpunct}{\mcitedefaultseppunct}\relax
\EndOfBibitem
\bibitem[Bruna \latin{et~al.}(2014)Bruna, Zaremba, Szlam, and
  LeCun]{gcn_iclr14}
Bruna,~J.; Zaremba,~W.; Szlam,~A.; LeCun,~Y. Spectral Networks and Locally
  Connected Networks on Graphs. International Conference on Learning
  Representations (ICLR). 2014\relax
\mciteBstWouldAddEndPuncttrue
\mciteSetBstMidEndSepPunct{\mcitedefaultmidpunct}
{\mcitedefaultendpunct}{\mcitedefaultseppunct}\relax
\EndOfBibitem
\bibitem[Duvenaud \latin{et~al.}(2015)Duvenaud, Maclaurin, Iparraguirre,
  Bombarell, Hirzel, Aspuru-Guzik, and Adams]{DuvenaudMAGHAA15}
Duvenaud,~D.~K.; Maclaurin,~D.; Iparraguirre,~J.; Bombarell,~R.; Hirzel,~T.;
  Aspuru-Guzik,~A.; Adams,~R.~P. \emph{Advances in Neural Information
  Processing Systems (NIPS) 28}; Curran Associates, Inc., 2015; pp
  2224--2232\relax
\mciteBstWouldAddEndPuncttrue
\mciteSetBstMidEndSepPunct{\mcitedefaultmidpunct}
{\mcitedefaultendpunct}{\mcitedefaultseppunct}\relax
\EndOfBibitem
\bibitem[Kearnes \latin{et~al.}(2016)Kearnes, McCloskey, Berndl, Pande, and
  Riley]{gcn_camd16}
Kearnes,~S.; McCloskey,~K.; Berndl,~M.; Pande,~V.; Riley,~P. Molecular graph
  convolutions: moving beyond fingerprints. \emph{Journal of Computer-Aided
  Molecular Design (CAMD)} \textbf{2016}, \emph{30}, 595--608\relax
\mciteBstWouldAddEndPuncttrue
\mciteSetBstMidEndSepPunct{\mcitedefaultmidpunct}
{\mcitedefaultendpunct}{\mcitedefaultseppunct}\relax
\EndOfBibitem
\bibitem[Gilmer \latin{et~al.}(2017)Gilmer, Schoenholz, Riley, Vinyals, and
  Dahl]{mpnn_icml17}
Gilmer,~J.; Schoenholz,~S.~S.; Riley,~P.~F.; Vinyals,~O.; Dahl,~G.~E. Neural
  Message Passing for Quantum Chemistry. Proceedings of the 34th International
  Conference on Machine Learning (ICML). 2017; pp 1263--1272\relax
\mciteBstWouldAddEndPuncttrue
\mciteSetBstMidEndSepPunct{\mcitedefaultmidpunct}
{\mcitedefaultendpunct}{\mcitedefaultseppunct}\relax
\EndOfBibitem
\bibitem[Altae-Tran \latin{et~al.}(2017)Altae-Tran, Ramsundar, Pappu, and
  Pande]{lddd_acs17}
Altae-Tran,~H.; Ramsundar,~B.; Pappu,~A.~S.; Pande,~V. Low Data Drug Discovery
  with One-Shot Learning. \emph{ACS Central Science} \textbf{2017}, \emph{3},
  283--293\relax
\mciteBstWouldAddEndPuncttrue
\mciteSetBstMidEndSepPunct{\mcitedefaultmidpunct}
{\mcitedefaultendpunct}{\mcitedefaultseppunct}\relax
\EndOfBibitem
\bibitem[Marcheggiani and Titov(2017)Marcheggiani, and Titov]{MarcheggianiT17}
Marcheggiani,~D.; Titov,~I. Encoding Sentences with Graph Convolutional
  Networks for Semantic Role Labeling. Proceedings of the 2017 Conference on
  Empirical Methods in Natural Language Processing. 2017; pp 1506--1515\relax
\mciteBstWouldAddEndPuncttrue
\mciteSetBstMidEndSepPunct{\mcitedefaultmidpunct}
{\mcitedefaultendpunct}{\mcitedefaultseppunct}\relax
\EndOfBibitem
\bibitem[Caruana(1997)]{Caruana1997}
Caruana,~R. Multitask Learning. \emph{Machine Learning} \textbf{1997},
  \emph{28}, 41--75\relax
\mciteBstWouldAddEndPuncttrue
\mciteSetBstMidEndSepPunct{\mcitedefaultmidpunct}
{\mcitedefaultendpunct}{\mcitedefaultseppunct}\relax
\EndOfBibitem
\bibitem[Ruder(2017)]{Ruder17a}
Ruder,~S. An Overview of Multi-Task Learning in Deep Neural Networks.
  \emph{CoRR} \textbf{2017}, \emph{abs/1706.05098}\relax
\mciteBstWouldAddEndPuncttrue
\mciteSetBstMidEndSepPunct{\mcitedefaultmidpunct}
{\mcitedefaultendpunct}{\mcitedefaultseppunct}\relax
\EndOfBibitem
\bibitem[Zhao~Chen and Rabinovich(2018)Zhao~Chen, and Rabinovich]{Gradnorm17}
Zhao~Chen,~C.-Y.~L.,~Vijay~Badrinarayanan; Rabinovich,~A. GradNorm: Gradient
  Normalization for Adaptive Loss Balancing in Deep Multitask Networks. ICML.
  2018\relax
\mciteBstWouldAddEndPuncttrue
\mciteSetBstMidEndSepPunct{\mcitedefaultmidpunct}
{\mcitedefaultendpunct}{\mcitedefaultseppunct}\relax
\EndOfBibitem
\bibitem[Kendall \latin{et~al.}(2018)Kendall, Gal, and Cipolla]{KendallGC17}
Kendall,~A.; Gal,~Y.; Cipolla,~R. Multi-Task Learning Using Uncertainty to
  Weigh Losses for Scene Geometry and Semantics. Proceedings of the IEEE
  Conference on Computer Vision and Pattern Recognition ({CVPR}). 2018\relax
\mciteBstWouldAddEndPuncttrue
\mciteSetBstMidEndSepPunct{\mcitedefaultmidpunct}
{\mcitedefaultendpunct}{\mcitedefaultseppunct}\relax
\EndOfBibitem
\bibitem[Rumelhart \latin{et~al.}(1988)Rumelhart, Hinton, and
  Williams]{Rumelhart1988}
Rumelhart,~D.~E.; Hinton,~G.~E.; Williams,~R.~J. In \emph{Neurocomputing:
  Foundations of Research}; Anderson,~J.~A., Rosenfeld,~E., Eds.; MIT Press:
  Cambridge, MA, USA, 1988; Chapter Learning Representations by
  Back-propagating Errors, pp 696--699\relax
\mciteBstWouldAddEndPuncttrue
\mciteSetBstMidEndSepPunct{\mcitedefaultmidpunct}
{\mcitedefaultendpunct}{\mcitedefaultseppunct}\relax
\EndOfBibitem
\bibitem[Jain \latin{et~al.}(2013)Jain, Ong, Hautier, Chen, Richards, Dacek,
  Cholia, Gunter, Skinner, Ceder, and Persson]{Jain2013}
Jain,~A.; Ong,~S.~P.; Hautier,~G.; Chen,~W.; Richards,~W.~D.; Dacek,~S.;
  Cholia,~S.; Gunter,~D.; Skinner,~D.; Ceder,~G.; Persson,~K.~a. {The Materials
  Project: A materials genome approach to accelerating materials innovation}.
  \emph{APL Materials} \textbf{2013}, \emph{1}, 011002\relax
\mciteBstWouldAddEndPuncttrue
\mciteSetBstMidEndSepPunct{\mcitedefaultmidpunct}
{\mcitedefaultendpunct}{\mcitedefaultseppunct}\relax
\EndOfBibitem
\bibitem[Xu \latin{et~al.}(2017)Xu, Ma, Liaw, Sheridan, and Svetnik]{Xu17}
Xu,~Y.; Ma,~J.; Liaw,~A.; Sheridan,~R.~P.; Svetnik,~V. Demystifying Multitask
  Deep Neural Networks for Quantitative Structure–Activity Relationships.
  \emph{Journal of Chemical Information and Modeling} \textbf{2017}, \emph{57},
  2490--2504, PMID: 28872869\relax
\mciteBstWouldAddEndPuncttrue
\mciteSetBstMidEndSepPunct{\mcitedefaultmidpunct}
{\mcitedefaultendpunct}{\mcitedefaultseppunct}\relax
\EndOfBibitem
\bibitem[Bingel and S{\o}gaard(2017)Bingel, and S{\o}gaard]{Bingel17}
Bingel,~J.; S{\o}gaard,~A. Identifying beneficial task relations for multi-task
  learning in deep neural networks. Proceedings of the 15th Conference of the
  European Chapter of the Association for Computational Linguistics: Volume 2,
  Short Papers. 2017; pp 164--169\relax
\mciteBstWouldAddEndPuncttrue
\mciteSetBstMidEndSepPunct{\mcitedefaultmidpunct}
{\mcitedefaultendpunct}{\mcitedefaultseppunct}\relax
\EndOfBibitem
\bibitem[Benesty \latin{et~al.}(2009)Benesty, Chen, Huang, and
  Cohen]{benesty2009pearson}
Benesty,~J.; Chen,~J.; Huang,~Y.; Cohen,~I. \emph{Noise reduction in speech
  processing}; Springer, 2009; pp 1--4\relax
\mciteBstWouldAddEndPuncttrue
\mciteSetBstMidEndSepPunct{\mcitedefaultmidpunct}
{\mcitedefaultendpunct}{\mcitedefaultseppunct}\relax
\EndOfBibitem
\bibitem[Myers and Well(2003)Myers, and Well]{myers2003research}
Myers,~J.; Well,~A. \emph{Research Design and Statistical Analysis}; Research
  Design and Statistical Analysis v. 1; Lawrence Erlbaum Associates, 2003\relax
\mciteBstWouldAddEndPuncttrue
\mciteSetBstMidEndSepPunct{\mcitedefaultmidpunct}
{\mcitedefaultendpunct}{\mcitedefaultseppunct}\relax
\EndOfBibitem
\bibitem[Kingma and Ba(2015)Kingma, and Ba]{KingmaB14}
Kingma,~D.~P.; Ba,~J. Adam: {A} Method for Stochastic Optimization.
  International Conference on Learning Representations (ICLR). 2015\relax
\mciteBstWouldAddEndPuncttrue
\mciteSetBstMidEndSepPunct{\mcitedefaultmidpunct}
{\mcitedefaultendpunct}{\mcitedefaultseppunct}\relax
\EndOfBibitem
\end{mcitethebibliography}

\end{document}